\documentclass[10pt,twocolumn,letterpaper]{article}

\usepackage{iccv}              

\usepackage{multirow}
\usepackage{xcolor}
\usepackage{colortbl}
\usepackage{graphicx} 
\definecolor{iccvblue}{rgb}{0.21,0.49,0.74}
\usepackage[pagebackref,breaklinks,colorlinks,allcolors=iccvblue]{hyperref}

\def\paperID{12653} 
\def\confName{ICCV}
\def\confYear{2025}

\title{ACAM-KD: Adaptive and Cooperative Attention Masking for Knowledge Distillation}

\author{
Qizhen Lan and Qing Tian\thanks{Corresponding author. This work was supported by the National Science Foundation (NSF) under Award No. 2153404 and No. 2412285.}\\
University of Alabama at Birmingham \\
{\tt\small \{qlan, qtian\}@uab.edu}
}

\begin{document}
\maketitle
\begin{abstract}
Dense visual prediction tasks, such as detection and segmentation, are crucial for time-critical applications (e.g., autonomous driving and video surveillance). While deep models achieve strong performance, their efficiency remains a challenge. Knowledge distillation (KD) is an effective model compression technique, but existing feature-based KD methods rely on static, teacher-driven feature selection, failing to adapt to the student's evolving learning state or leverage dynamic student-teacher interactions. To address these limitations, we propose Adaptive student-teacher Cooperative Attention Masking for Knowledge Distillation (ACAM-KD), which introduces two key components: (1) Student-Teacher Cross-Attention Feature Fusion (STCA-FF), which adaptively integrates features from both models for a more interactive distillation process, and (2) Adaptive Spatial-Channel Masking (ASCM), which dynamically generates importance masks to enhance both spatial and channel-wise feature selection. Unlike conventional KD methods, ACAM-KD adapts to the student's evolving needs throughout the entire distillation process. Extensive experiments on multiple benchmarks validate its effectiveness. For instance, on COCO2017, ACAM-KD improves object detection performance by up to 1.4 mAP over the state-of-the-art when distilling a ResNet-50 student from a ResNet-101 teacher. For semantic segmentation on Cityscapes, it boosts mIoU by 3.09 over the baseline with DeepLabV3-MobileNetV2 as the student model.
\end{abstract}    
\section{Introduction}
\label{sec:intro}
Deep learning models have significantly advanced dense visual prediction tasks, such as object detection and segmentation, leading to breakthroughs in applications like autonomous driving, surveillance, and robotics. Despite these successes, deploying high-performance deep models on resource-limited devices remains challenging due to their high computational complexity and large parameter sizes. 
Knowledge distillation (KD) has emerged as an effective solution, transferring knowledge from a high-capacity teacher model to a lightweight student model to maintain accuracy while reducing complexity. Among various KD approaches, feature-based knowledge distillation is particularly well-suited for dense prediction tasks like object detection and segmentation. By transferring intermediate feature representations instead of final predictions, it effectively preserves the dense spatial information essential for these tasks.

\begin{figure}[t]
    \centering
    \includegraphics[width=\columnwidth]{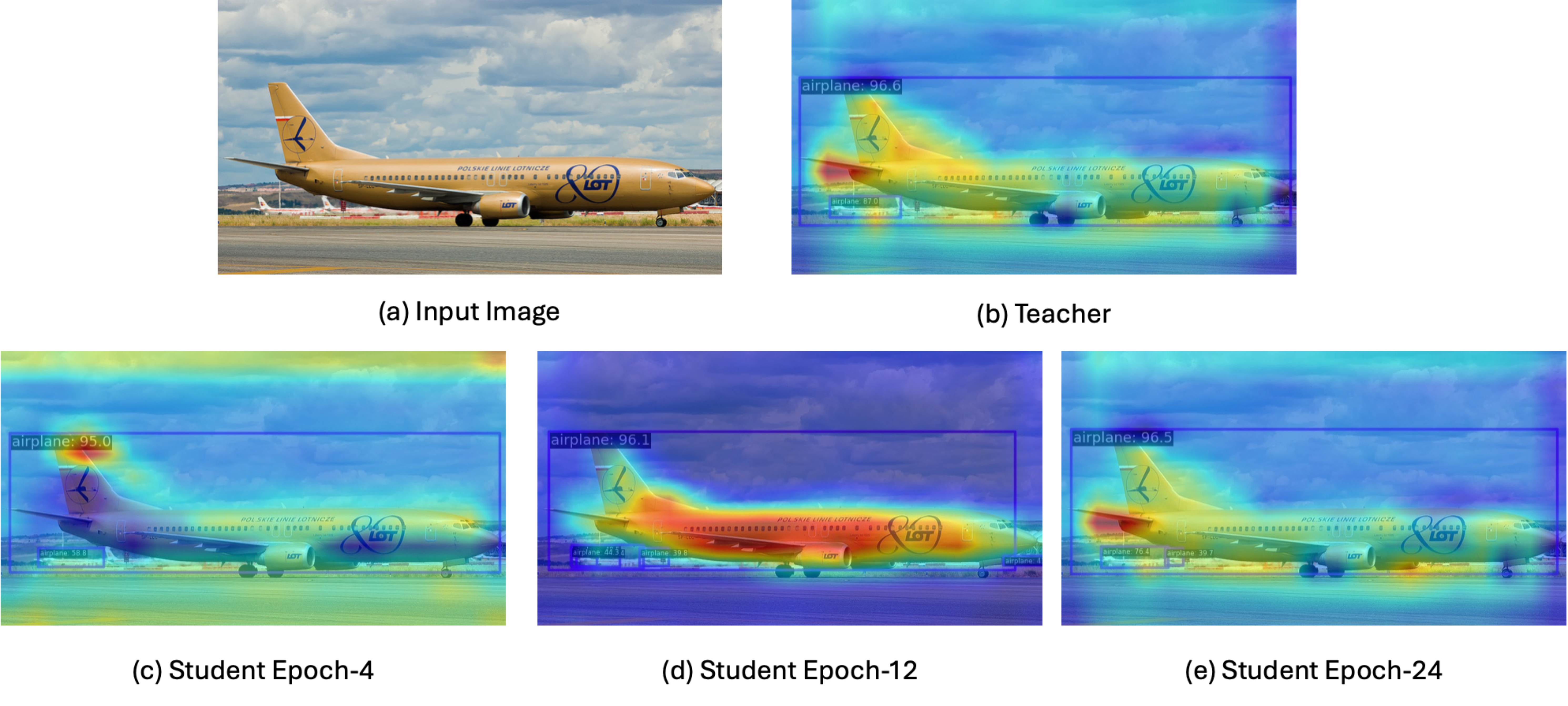}
\caption{
    Visualization of attention maps from different models at various training stages. (a) input image, (b) teacher model’s attention, and (c-e) student model's attention at epochs 4, 12, and 24. The color represents attention intensity, with red indicating the highest focus and blue the lowest. The teacher model’s attention is static and suboptimal. The student’s attention evolves over time, with epoch 12 demonstrating better localization than the teacher.
}
    \label{fig:attention_maps}
\end{figure}

However, most existing feature-based KD methods rely on fixed or teacher-driven feature selection. They assume that the most critical regions for distillation can be determined solely by the teacher \cite{zhang2020improve, huang2023masked, zhang2024freekd} or predefined heuristics, such as bounding boxes, RPN regions, prediction confidence \cite{yang2022focal, li2017mimicking, yang2022prediction}.
These methods require the student to passively focus on predetermined regions, disregarding its evolving learning state and unique characteristics. However, the features deemed important by the pre-trained teacher or human heuristics may not align with what the student actually needs at different learning stages.
Moreover, in these approaches, the focused areas for distillation remain static for a given image, failing to adapt to the student’s dynamically changing knowledge (i.e., feature representations). Repeatedly emphasizing the same regions in every epoch is inefficient, especially when the student has already acquired sufficient understanding of them. 
To make things worse, an inaccurate teacher focus likely misleads the student. Fig. \ref{fig:attention_maps} provides an example. The student initially exhibits inaccurate attention (Epoch 4) but improves by Epoch 12, focusing better on the foreground plane. However, by Epoch 24, the student’s attention shifts to resemble the teacher’s fixed feature selection instead of further refining its own learning.
In addition, while existing methods primarily focus on spatial feature selection—an essential aspect of dense prediction tasks—they often overlook the significance of channel-wise feature selection and fail to consider the varying contributions of different channels.

To address the above challenges in feature-based knowledge distillation, we propose Adaptive Student-Teacher Cooperative Attention Masking for enhanced feature-based knowledge distillation (ACAM-KD). Specifically, ACAM-KD contains two key modules:
(1) Student-Teacher Cross-Attention Feature Fusion (STCA-FF) module, which integrates features from both the pre-trained teacher and the dynamically evolving student models.
(2) Building on the cross-attention features from STCA-FF, the second module, Adaptive Spatial-Channel Masking (ASCM), adaptively generates importance masks across both spatial and channel dimensions for more effective distillation.
Unlike traditional methods that rely solely on fixed or teacher-defined attention, ACAM-KD leverages cross-attention to facilitate a cooperative knowledge transfer process, allowing the student to dynamically interact with the teacher and refine feature selection based on both teacher-provided knowledge and the student's own evolving learning state (i.e., feature representations).
The attention masks for distillation are dynamically updated throughout the student’s learning process, enabling the distillation to adaptively focus on different tensor regions as the student evolves and its learning needs change.
We validate the effectiveness of our approach through extensive experiments on various dense prediction tasks, including object detection and segmentation. Our method consistently outperforms state-of-the-art KD approaches across multiple benchmarks, demonstrating superior performance and adaptability in student models distilled with our framework.

\section{Related Work}
\label{sec:re_work}

\subsection{Knowledge Distillation for Object Detection}
A key issue in KD for object detection is to identify which spatial regions and feature representations are most important. FGFI \cite{wang2019distilling} introduced pixel-level masks to highlight regions near ground-truth boxes, while GID \cite{dai2021general} emphasized areas where teacher and student predictions differ significantly. FKD \cite{zhang2020improve} leveraged high-attention features to weigh important regions, and FGD \cite{yang2022focal} further advanced this by also incorporating ground-truth bounding boxes and capturing global context. Recent works explore learnable feature selection strategies. MasKD \cite{huang2023masked} proposed learnable receptive tokens, derived from the teacher’s features, to generate pixel-wise masks that capture fine-grained spatial knowledge. FreeKD \cite{zhang2024freekd} introduced a frequency-domain semantic prompt to guide the student in focusing on essential feature regions, improving distillation robustness in dense prediction tasks. Despite these advancements, existing KD methods for object detection remain predominantly heuristics-based or teacher-driven, assuming static feature importance throughout training. 
Although \cite{huang2023masked} and \cite{zhang2024freekd} attempt to learn tokens or prompts for masking, their learning process is conducted offline in one shot, relying solely on a fixed pre-trained teacher. Consequently, the learned tokens or prompts remain static throughout the student's learning process, failing to account for the student's evolving state and its dynamic interaction with the teacher model. As a result, for a given image, the student consistently receives the same knowledge across epochs, regardless of its progress.
Also, the dynamic interplay between spatial and channel-wise feature selection is often neglected.

\subsection{Knowledge Distillation for Semantic Segmentation}
Effective knowledge distillation for semantic segmentation is challenging due to high-resolution outputs and complex pixel dependencies. Early KD methods primarily addressed this by enforcing local spatial consistency, aligning teacher and student features at the pixel level. SKD \cite{liu2019structured} introduced pairwise and holistic distillation strategies, with the former preserving local pixel relationships through pairwise similarity alignment and the latter enforcing high-order structural consistency via adversarial training on score maps. IFVD \cite{wang2020intra} enhanced knowledge transfer by enforcing intra-class feature consistency, improving the stability of learned representations across spatial regions. CWD \cite{shu2021channel} introduced a channel-wise feature alignment strategy to help the student model learn the teacher’s feature distribution across channels. CIRKD \cite{yang2022cross} further advanced relational distillation by modeling both intra-image and cross-image dependencies, promoting structured feature learning beyond direct pixel-wise supervision.

Although progress has been made, existing KD approaches for segmentation share similar limitations with KD methods for detection, relying on heuristics or teacher-driven, fixed feature importance that fails to adapt to the student's evolving state. Additionally, they lack a joint optimization of spatial and channel-wise knowledge selection. To address these shortcomings, we propose Adaptive Student-Teacher Cooperative Attention Masking for enhanced feature-based knowledge distillation (ACAM-KD). The following section details our methodology.

\begin{figure*}[t]
    \centering
    \includegraphics[width=0.99\textwidth]{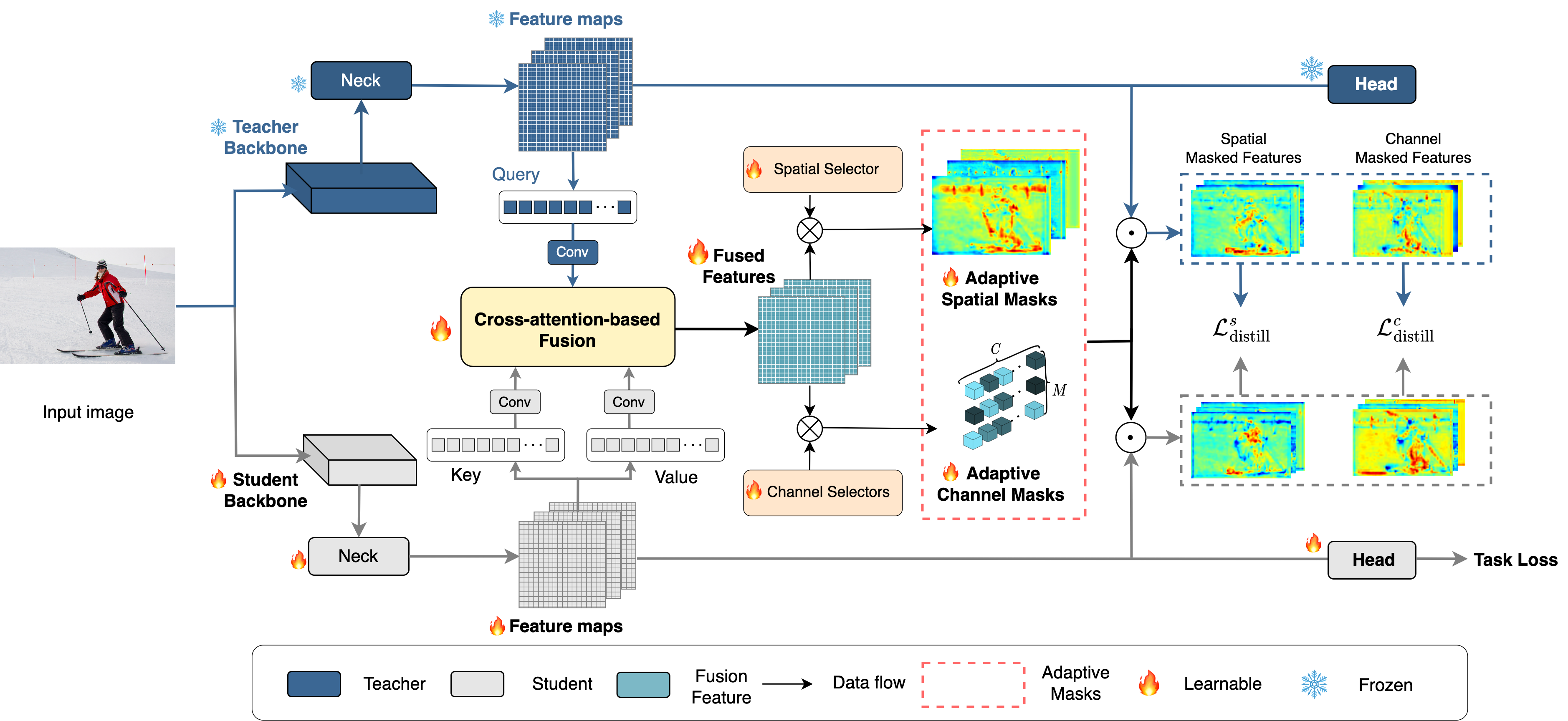}
\caption{Overview of the proposed Adaptive and Cooperative Attention Masking for Knowledge Distillation (ACAM-KD) framework. ACAM-KD leverages student-teacher cross-attention to dynamically and cooperatively guide the distillation process. Instead of relying on static, teacher-determined attention masking, our distillation masks are updated throughout distillation, adapting to the student's evolving learning needs and selectively focusing on the most beneficial spatial and channel-wise features at a particular learning stage. Different colors in the adaptive channel masks indicate varying levels of importance.}
\label{fig:main_figure}
\end{figure*}

\section{Methodology}
\label{sec:method}
Knowledge distillation (KD) transfers knowledge from a high-capacity teacher model to a lightweight student model, aiming to improve efficiency while maintaining performance. Given a teacher model $T$ and a student model $S$, feature-based KD minimizes the distillation loss $\mathcal{L}_{\text{feat}}$:
\begin{equation}
    \frac{1}{M} \sum_{m=1}^{M} 
    \frac{1}{C \sum_{p=1}^{H\times W}\mathbf{M}_{m,p}}
    \Big\|
    \mathbf{M}_m \odot (F^T 
    -  f_{\text{align}}(F^S)) 
    \Big\|_2^2.
\end{equation}
\noindent where $F^{T}$ and $F^{S}$ are the teacher’s and student’s feature maps, respectively. The function $f_{\text{align}}$ is an adaptation layer applied to $F^{S}$ to align its feature dimension with $F^{T}$. The mask $\mathbf{M}_m$ is applied element-wise to emphasize relevant regions for distillation, where $\mathbf{M}_{m,p}$ represents the selection weight at spatial position $p$ of the m-th mask. $C$, $H$, and $W$ denote the number of channels, feature map height, and width, respectively. The loss is averaged over $M$ masks.

Existing feature-based KD methods rely on fixed, teacher-driven mask selection, assuming key regions can be identified solely from teacher activations or predefined heuristics. This approach disregards the student’s role and evolving knowledge, failing to adapt to its dynamically changing needs at different learning stages. Additionally, maintaining static focus regions throughout distillation is inefficient or even dangerous, and most methods emphasize spatial selection while neglecting channel-wise importance. To address these limitations, we propose Adaptive Student-Teacher Cooperative Attention Masking for Knowledge Distillation (ACAM-KD), which dynamically refines feature selection through student-teacher interactions. Instead of relying on predefined masks, our approach learns masks adaptively and continuously throughout the KD process. \cref{fig:main_figure} presents an overview of ACAM-KD, while Sec. \ref{sec:stca-ff} and \ref{sec:ascm} detail its two main components.

\subsection{Student-Teacher Cross-Attention Feature Fusion for Distillation Attention}\label{sec:stca-ff}

In knowledge distillation, the student should not only passively absorb knowledge from the teacher but also identify the most beneficial learning regions based on its evolving needs. To enable this, we first introduce a student-teacher cross-attention-based feature fusion mechanism (STCA-FF), where the teacher’s feature map defines the query, while the student’s feature map provides both the key and value. 
Given teacher and student feature maps $F^T, F^S \in \mathbb{R}^{C \times H \times W}$, we apply $1 \times 1$ convolutions to project them into query, key, and value representations:
\begin{equation}
    Q = W_q F^T, \quad K = W_k F^S, \quad V = W_v F^S,
\end{equation}
\noindent where $W_q \in \mathbb{R}^{C_q \times C}$ and $W_k \in \mathbb{R}^{C_q \times C}$ reduce the channel dimension to $C_q = C/2$, while $W_v \in \mathbb{R}^{C \times C}$ maintains the original channel dimension.
To compute attention over spatial locations, the feature maps are flattened along the spatial dimensions $(H, W)$, yielding $Q \in \mathbb{R}^{H W \times C_q}$, $K \in \mathbb{R}^{C_q \times H W}$, and $V \in \mathbb{R}^{H W \times C}$. The attention matrix is then computed as:
\begin{equation}
    A = \text{softmax} \left( \frac{Q K}{\sqrt{C_q}} \right) \in \mathbb{R}^{H W \times H W}.
\end{equation}
Using the attention weights, we compute the fused features:
\begin{equation}
    F_{\text{fused}} = A V \in \mathbb{R}^{C \times H \times W}.
\end{equation}

The fused features integrate information from both the teacher and student, enabling the distillation process to leverage their joint interactions rather than relying solely on predefined teacher activations. This fused representation is then fed into the next adaptive spatial-channel masking module, where learnable selection units generate spatial and channel-wise masks to prioritize regions for distillation, ensuring knowledge transfer is dynamically adjusted based on both teacher and student features.

\subsection{Adaptive Spatial-Channel Masking}\label{sec:ascm}

Unlike frozen teacher features, the student-teacher interacted features dynamically evolve throughout the student's learning process. Therefore, the distillation mask applied to these fused features should also adapt accordingly. However, the fixed attention masks commonly used in existing methods fail to accommodate this need. To overcome this limitation, we introduce a learnable selection mechanism that dynamically generates feature importance masks during distillation, enabling the student model to selectively focus on the most beneficial spatial and channel-wise regions at a certain learning stage.

Unlike existing spatial-only masking methods, we introduce two complementary sets of learnable selection units for distillation masking: channel-wise selection units $m^{c} \in \mathbb{R}^{M}$ and spatial selection units $m^c \in \mathbb{R}^{M \times C}$. These learnable selection units operate on the dynamically evolving student-teacher fused features, generating adaptive channel-wise and spatial masks to enhance distillation. Given the fused feature representation $F_{\text{fused}} \in \mathbb{R}^{C \times H \times W}$, the selection units generate channel and spatial masks as follows:
\begin{equation}
    \mathbf{M}^{c} = \sigma(m^c \cdot v), \quad \mathbf{M}^s = \sigma(m^s \cdot z),
\end{equation}
\noindent where $\sigma(\cdot)$ is the sigmoid function, $v \in \mathbb{R}^{1 \times C}$ is the spatially average-pooled vector of $F_{\text{fused}}$, and $z \in \mathbb{R}^{C \times H W}$ is the flattened $F_{\text{fused}}$ along the spatial dimension.
With the masks, we define the channel-wise and spatial feature distillation losses $\mathcal{L}_{\text{distill}}^{c}$ and $\mathcal{L}_{\text{distill}}^s$ as:

\vspace{-0.05in}
\fontsize{8.25pt}{10pt}\selectfont
\begin{equation}
    \mathcal{L}_{\text{distill}}^{c} = \frac{1}{M} \sum_{m=1}^{M} 
    \frac{1}{HW\sum_{k=1}^{C}\mathbf{M}^{c}_{m,k}}
    \Big\|
    \mathbf{M}^{c}_m \odot (F^T 
    - f_{\text{align}}(F^S)) 
    \Big\|_2^2,
\end{equation}
\begin{equation}
    \mathcal{L}_{\text{distill}}^s = \frac{1}{M} \sum_{m=1}^{M} 
    \frac{1}{C\sum_{p=1}^{H\times W}\mathbf{M}^s_{m,p}}
    \Big\|
    \mathbf{M}^s_m \odot (F^T 
    - f_{\text{align}}(F^S)) 
    \Big\|_2^2,
\end{equation}
\normalsize

\noindent where $k$ indicates the index of channels. By optimizing the two losses, the adaptive masks $ \mathbf{M}^{c} \in \mathbb{R}^{M \times C}$ and $ \mathbf{M}^s \in \mathbb{R}^{M \times H \times W} $ dynamically adjust to emphasize channels and spatial locations that are important from the teacher’s perspective while adapting to the student's evolving needs at each distillation stage. This dual awareness arises from $F_{\text{fused}}$, which represents the collaborative outcome of both the teacher and student.



\paragraph{Mask Diversity}
To prevent trivial solutions where all masks converge to similar patterns or degenerate into zero values, we incorporate a Dice coefficient-based diversity loss that promotes variation by penalizing excessive similarity among masks. Given a set of learnable masks $\{M_i\}_{i=1}^{M}$, we define the diversity loss as:

\begin{equation}
    \mathcal{L}_{\text{div}} = \frac{2 \sum_{i=1}^{M} \sum_{\substack{j=1 \\ j \neq i}}^{M} M_i \cdot M_j}
    {\sum_{i=1}^{M} M_i^2 + \sum_{j=1}^{M} M_j^2},
\end{equation}

\noindent 
where the numerator measures the pairwise similarity between different masks, while the denominator normalizes it by the sum of the squared magnitudes of the masks. 

\subsection{Overall Loss}

The total training objective is the sum of three losses:
\begin{equation}
    \mathcal{L} = \mathcal{L}_{\text{task}} + \alpha (\mathcal{L}_{\text{distill}}^c + \mathcal{L}_{\text{distill}}^s) + \lambda \mathcal{L}_{\text{div}}.
\end{equation}
\noindent where $\mathcal{L}_{\text{task}}$ is the task (e.g., detection or segmentation) loss, $\mathcal{L}_{\text{distill}}^s$ and $\mathcal{L}_{\text{distill}}^c$ are the spatial and channel-wise distillation losses, and $\mathcal{L}_{\text{div}}$ is the mask diversity loss. 
$\alpha$ and $\lambda$ are balancing hyperparameters. In our experiments, we set $\alpha = 1$ and $\lambda = 1$, which already achieves strong performance. 

\section{Experiments and Results}
\label{sec:experiments}

To demonstrate the effectiveness of our ACAM-KD, we conduct extensive experiments across diverse model architectures on both object detection and semantic segmentation tasks.

\subsection{Object Detection}

\subsubsection{Experimental Setup}
We conduct experiments on the widely used Microsoft COCO2017 detection dataset \cite{lin2014microsoft}, which includes 118,287 training images and 5,000 validation images from 80 object classes. 
We evaluate ACAM-KD on various detectors, including one-stage \cite{lin2017focal}, two-stage \cite{Ren_2017}, and anchor-free detectors \cite{yang2019reppoints}.
We perform feature distillation at the FPN neck, implementing our approach within the MMDetection framework \cite{mmdetection} and following its official training strategies. All models are optimized using SGD with a momentum of 0.9 and a weight decay of 0.0001. In all experiments, we set $M$ to 6 for both spatial and channel cases. Additionally, we adopt the inheritance strategy from privious knowledge distillation works \cite{huang2023knowledge, yang2022masked} to stabilize early-stage training.

\subsubsection{Experimental Results}
\begin{table}[t]
    \centering
    \resizebox{\columnwidth}{!}{%
    \begin{tabular}{lcccccc}
        \toprule
        \textbf{Method} & \textbf{mAP} & \textbf{AP$_{50}$} & \textbf{AP$_{75}$} & \textbf{AP$_s$} & \textbf{AP$_m$} & \textbf{AP$_l$} \\
        \midrule
        \multicolumn{7}{c}{\textit{Single-Stage Detectors (RetinaNet)}} \\
        \midrule
        \textbf{T: R101} & 38.9 & 58.0 & 41.5 & 21.0 & 42.8 & 52.4 \\
        \textbf{S: R50}  & 37.4 & 56.7 & 39.6 & 20.0 & 40.7 & 49.7 \\
        FitNet \cite{romero2014fitnets} & 37.4 & 57.1 & 40.0 & 20.8 & 40.8 & 50.9 \\
        GID \cite{dai2021general}     & 39.1 & 59.0 & 42.3 & 22.8 & 43.1 & 52.3 \\
        FRS \cite{zhixing2021distilling}     & 39.3 & 58.8 & 42.0 & 21.5 & 43.3 & 52.6 \\
        FGD \cite{yang2022focal}     & 39.6 & -    & -    & 22.9 & 43.7 & 53.6 \\
        MasKD \cite{huang2023masked}   & 39.8 & 59.0 & 42.5 & 21.5 & 43.9 & 54.0 \\
        FreeKD \cite{zhang2024freekd}   & 39.9 & -    & -    & 21.2 & 44.0 & 53.7 \\
        \rowcolor{gray!20} \textbf{Ours}    & \textbf{41.2} & \textbf{60.6} & \textbf{44.1} & \textbf{24.6} & \textbf{45.5} & \textbf{54.1} \\
        \midrule
        \multicolumn{7}{c}{\textit{Two-Stage Detectors (Faster R-CNN)}} \\
        \midrule
        \textbf{T: R101} & 39.8 & 60.1 & 43.3 & 22.5 & 43.6 & 52.8 \\
        \textbf{S: R50}  & 38.4 & 59.0 & 42.0 & 21.5 & 42.1 & 50.3 \\
        FitNet \cite{romero2014fitnets}  & 38.9 & 59.5 & 42.4 & 21.9 & 42.2 & 51.6 \\
        GID \cite{dai2021general}     & 40.2 & 60.7 & 43.8 & 22.7 & 44.0 & 53.2 \\
        FRS \cite{zhixing2021distilling}     & 39.5 & 60.1 & 43.3 & 22.3 & 43.6 & 51.7 \\
        FGD \cite{yang2022focal}     & 40.4 & -    & -    & 22.8 & 44.5 & 53.5 \\
        MasKD \cite{huang2023masked}   & 40.8 & 60.7 & 44.4 & 23.2 & 44.6 & 53.6 \\
        FreeKD \cite{zhang2024freekd}   & 40.8 & -    & -    & 23.1 & 44.7 & 54.0 \\
        \rowcolor{gray!20} \textbf{Ours}    & \textbf{41.4} & \textbf{61.7} & \textbf{45.2} & \textbf{24.7} & \textbf{45.0} & \textbf{53.3} \\
        \midrule
        \multicolumn{7}{c}{\textit{Anchor-Free Detectors (RepPoints)}} \\
        \midrule
        \textbf{T: R101} & 40.5 & 61.3 & 43.5 & 23.4 & 44.7 & 53.2 \\
        \textbf{S: R50}  & 38.6 & 59.6 & 41.6 & 22.5 & 42.2 & 50.4 \\
        FitNet \cite{romero2014fitnets}  & 40.7 & 61.6 & 43.9 & 23.7 & 44.4 & 53.3 \\
        FGD \cite{yang2022focal}     & 41.0 & -    & -    & 23.8 & 45.3 & 53.6 \\
        MasKD \cite{huang2023masked}   & 41.1 & 61.4 & 44.5 & 24.1 & 45.4 & 53.8 \\
        \rowcolor{gray!20} \textbf{Ours}    & \textbf{42.5} & \textbf{63.5} & \textbf{46.0} & \textbf{26.1} & \textbf{46.8} & \textbf{54.2} \\
        \bottomrule
    \end{tabular}
    }
    \caption{KD performance with ResNet-backboned teachers on COCO validation set. T: teacher model, S: student baseline.}
    \label{tab:detectors}
\end{table}
\begin{table}[h]
    \centering
    \resizebox{\columnwidth}{!}{%
    \begin{tabular}{lcccccc}
        \toprule
        \textbf{Method} & \textbf{mAP} & \textbf{AP$_{50}$} & \textbf{AP$_{75}$} & \textbf{AP$_s$} & \textbf{AP$_m$} & \textbf{AP$_l$} \\
        \midrule
        \multicolumn{7}{c}{\textit{Single-Stage Detectors (RetinaNet)}} \\
        \midrule
        \textbf{T: X101}     & 41.2 & 62.1 & 45.1 & 24.0 & 45.5 & 53.5 \\
        \textbf{S: R50}      & 37.4 & 56.7 & 39.6 & 20.0 & 40.7 & 49.7 \\
        COFD \cite{heo2019comprehensive}    & 37.8 & 58.3 & 41.1 & 21.6 & 41.2 & 48.3 \\
        FKD  \cite{zhang2020improve}    & 39.6 & 58.8 & 42.1 & 22.7 & 43.3 & 52.5 \\
        FRS \cite{zhixing2021distilling}    & 40.1 & 59.5 & 42.5 & 21.9 & 43.7 & 54.3 \\
        FGD \cite{yang2022focal}     & 40.4 & -    & -    & 23.4 & 44.7 & 54.1 \\
        MasKD \cite{huang2023masked}   & 40.9 & 60.1 & 43.6 & 22.8 & 45.3 & 55.1 \\
        FreeKD \cite{zhang2024freekd}  & 41.0 & -    & -    & 22.3 & 45.1 & 55.7 \\
        \rowcolor{gray!20} \textbf{Ours}    & \textbf{41.5} & \textbf{61.3} & \textbf{44.5} & \textbf{25.3} & \textbf{45.5} & \textbf{54.3} \\
        \midrule
        \multicolumn{7}{c}{\textit{Two-Stage (S: Faster R-CNN, T: Cascade Mask R-CNN)}} \\
        \midrule
        \textbf{T: X101} & 45.6 & 64.1 & 49.7 & 26.2 & 49.6 & 60.0 \\
        \textbf{S: R50}      & 38.4 & 59.0 & 42.0 & 21.5 & 42.1 & 50.3 \\
        LED \cite{chen2017learning}      & 38.7 & 59.0 & 42.1 & 22.0 & 41.9 & 51.0 \\
        FGFI \cite{wang2019distilling}     & 39.1 & 59.8 & 42.8 & 22.2 & 42.9 & 51.1 \\
        COFD \cite{heo2019comprehensive}     & 38.9 & 60.1 & 42.6 & 21.8 & 42.7 & 50.7 \\
        FKD \cite{zhang2020improve}      & 41.5 & 62.2 & 45.1 & 23.5 & 45.0 & 55.3 \\
        FGD \cite{yang2022focal}      & 42.0 & -    & -    & 23.7 & 46.4 & 55.5 \\
        MasKD \cite{huang2023masked}    & 42.4 & 62.9 & 46.8 & 24.2 & 46.7 & 55.9 \\
        FreeKD \cite{zhang2024freekd}  & 42.4 & -    & -    & 24.1 & 46.7 & 55.9 \\
        \rowcolor{gray!20} \textbf{Ours}    & \textbf{42.6} & \textbf{63.0} & \textbf{46.6} & \textbf{25.2} & \textbf{46.8} & \textbf{55.6} \\
        \midrule
        \multicolumn{7}{c}{\textit{Anchor-Free Detectors (RepPoints)}} \\
        \midrule
        \textbf{T: X101}     & 44.2 & 65.5 & 47.8 & 26.2 & 48.4 & 58.5 \\
        \textbf{S: R50}      & 38.6 & 59.6 & 41.6 & 22.5 & 42.2 & 50.4 \\
        FKD \cite{zhang2020improve}     & 40.6 & 61.7 & 43.8 & 23.4 & 44.6 & 53.0 \\
        FGD \cite{yang2022focal}     & 41.3 & -    & -    & 24.5 & 45.2 & 54.0 \\
        MasKD \cite{huang2023masked}   & 41.8 & 62.6 & 45.1 & 24.2 & 45.4 & 55.2 \\
        FreeKD \cite{zhang2024freekd}  & 42.4 & -    & -    & 24.3 & 46.4 & 56.6 \\
        \rowcolor{gray!20} \textbf{Ours}    & \textbf{42.8} & \textbf{63.9} & \textbf{46.4} & \textbf{26.7} & \textbf{47.1} & \textbf{54.5} \\
        \bottomrule
    \end{tabular}
    }
    \caption{KD performance with ResNeXt-backboned teachers on COCO validation set. T: teacher model, S: student baseline.}
    \label{tab:detectors_strong}
\end{table}

\cref{tab:detectors} and \cref{tab:detectors_strong} present object detection performance on COCO under different teacher configurations. With a ResNet-50 (R50) student, our method consistently surpasses previous distillation approaches across single-stage (RetinaNet \cite{lin2017focal}), two-stage (Faster R-CNN \cite{Ren_2017}), and anchor-free (RepPoints \cite{yang2019reppoints}) detectors. When distilled from a ResNet-101 (R101) teacher, our method improves the student's mAP by 1.3 for single-stage, 0.6 for two-stage, and 1.4 for anchor-free detectors over the second-best KD method. With a ResNeXt-101 (X101) teacher, we achieve even greater gains, boosting the student's mAP by 4.1 for single-stage, 4.2 for two-stage, and 4.2 for anchor-free detectors.
These results demonstrate our method's effectiveness in improving student detectors' efficiency and performance.

\subsection{Segmentation}

\subsubsection{Experimental Setup}
We validate our method using the Cityscapes dataset \cite{cordts2016cityscapes}, which contains a total of 5,000 finely annotated images and 19,998 coarsely annotated images, captured from 50 cities in different weather conditions and seasons. Like most previous works in this area, we only use the finely annotated images. We evaluate all segmentors using mean Intersection-over-Union (mIoU). In all experiments, we adopt the DeepLabV3 framework \cite{chen2018encoder} with a ResNet-101 backbone as the teacher network. For student models, we explore various configurations, combining frameworks such as DeepLabV3 and PSPNet \cite{zhao2017pyramid} with backbones like ResNet-18 and MobileNetV2 \cite{sandler2018mobilenetv2}.

For the semantic segmentation task, we distill features from the predicted segmentation maps. The models are trained using the MMSegmentation \cite{mmseg2020} framework with a 40K-iteration schedule and an input resolution of 512×1024. Optimization is performed using SGD with a weight decay of 5e-4, while a polynomial annealing scheduler is applied to adjust the learning rate, starting at 0.02. To ensure that each learnable mask corresponds to a specific semantic category, we set $M$ to 19, matching the number of classes in the Cityscapes dataset.

\subsubsection{Experimental Results}
\cref{tab:segmentation}, \cref{tab:deeplabv3_mbv2}, and \cref{tab:pspnet_r18} present the semantic segmentation performance across different student architectures on the Cityscapes dataset. FLOPs are calculated based on an input size of $512\times1024$. As shown, our method consistently outperforms previous knowledge distillation approaches when distilling from a DeepLabV3-R101 teacher to various lightweight student models, including DeepLabV3 with ResNet-18 and MobileNetV2 \cite{sandler2018mobilenetv2} backbones. Furthermore, we investigate heterogeneous distillation by transferring knowledge from DeepLabV3-R101 to PSPNet \cite{zhao2017pyramid} with a ResNet-18 backbone.

For DeepLabV3-MBV2 as the student, our method improves mIoU by 3.09 over the baseline and 0.79 over the best competing KD method. With DeepLabV3-R18, our approach achieves 77.53 mIoU, surpassing the strongest competitor by 0.53 mIoU. Similarly, for PSPNet-R18, our method boosts mIoU by 3.44 over the baseline and 0.65 over the best prior method. These results underscore the effectiveness and generalizability of our approach across different segmentation models, consistently improving student performance in low-computation settings.
\begin{table}[h]
    \centering
    \resizebox{0.9\columnwidth}{!}{%
    \begin{tabular}{p{2cm} p{1.5cm} |p{0.8cm} p{0.8cm} p{0.8cm}}
        \toprule
        \multicolumn{2}{c|}{\textbf{Method}} & \textbf{Params (M)} & \textbf{FLOPs (G)} & \textbf{mIoU (\%)} \\
        \midrule
        \multicolumn{2}{l|}{\textbf{T: DeepLabV3-R101}} & \centering 84.74 & \centering 695 & 78.07 \\
        \midrule
        \multicolumn{2}{l|}{\textbf{S: DeepLabV3-R18}} & \centering \multirow{9}{*}{13.84} & \centering \multirow{9}{*}{120} & 72.96 \\
        SKD~\cite{liu2019structured} & \textcolor{gray}{CVPR'19} & & & 75.42 \\
        IFVD~\cite{wang2020intra} & \textcolor{gray}{ECCV'20} & & & 75.59 \\
        CWD~\cite{shu2021channel} & \textcolor{gray}{ICCV'21} & & & 75.55 \\
        CIRKD~\cite{yang2022cross} & \textcolor{gray}{CVPR'22} & & & 76.38 \\
        MasKD~\cite{huang2023masked} & \textcolor{gray}{ICLR'23} & & & 77.00 \\
        FreeKD~\cite{zhang2024freekd} & \textcolor{gray}{CVPR'24} & & & 76.45 \\
        MGD~\cite{yang2022masked} & \textcolor{gray}{ECCV'22} & & & 76.02 \\
        \rowcolor{gray!20} \textbf{Ours} & & & & \textbf{77.53} \\
        \bottomrule
    \end{tabular}
    }
    \caption{Comparison of knowledge distillation methods for semantic segmentation on Cityscapes with the DeepLabV3-R18 student model. T: teacher model, S: student baseline model.}
    \label{tab:segmentation}
\end{table}

\begin{table}[h]
    \centering
    \resizebox{0.9\columnwidth}{!}{%
    \begin{tabular}{p{2cm} p{1.5cm} |p{0.8cm} p{0.8cm} p{0.8cm}}
        \toprule
        \multicolumn{2}{c|}{\textbf{Method}} & \textbf{Params (M)} & \textbf{FLOPs (G)} & \textbf{mIoU (\%)} \\
        \midrule
        \multicolumn{2}{l|}{\textbf{T: DeepLabV3-R101}} & \centering 84.74 & \centering 695 & 78.07 \\
        \midrule
        \multicolumn{2}{l|}{\textbf{S: DeepLabV3-MBV2}} & \centering \multirow{6}{*}{3.2} & \centering \multirow{6}{*}{31.14} & 73.12 \\
        SKD~\cite{liu2019structured} & \textcolor{gray}{CVPR'19} & & & 73.82 \\
        IFVD~\cite{wang2020intra} & \textcolor{gray}{ECCV'20} & & & 73.50 \\
        CWD~\cite{shu2021channel} & \textcolor{gray}{ICCV'21} & & & 74.66 \\
        CIRKD~\cite{yang2022cross} & \textcolor{gray}{CVPR'22} & & & 75.42 \\
        MasKD~\cite{huang2023masked} & \textcolor{gray}{ICLR'23} & & & 75.26 \\
        \rowcolor{gray!20} \textbf{Ours} & & & & \textbf{76.21} \\
        \bottomrule
    \end{tabular}
    }
    \caption{Comparison of knowledge distillation methods for semantic segmentation on Cityscapes with the DeepLabV3-MBV2 student model. T: teacher model, S: student baseline model.}
    \label{tab:deeplabv3_mbv2}
\end{table}

\begin{table}[h]
    \resizebox{0.9\columnwidth}{!}{%
    \begin{tabular}{p{2cm} p{1.5cm} |p{0.8cm} p{0.8cm} p{0.8cm}}
        \toprule
        \multicolumn{2}{c|}{\textbf{Method}} & \textbf{Params (M)} & \textbf{FLOPs (G)} & \textbf{mIoU (\%)} \\
        \midrule
        \multicolumn{2}{l|}{\textbf{T: DeepLabV3-R101}} & \centering 84.74 & \centering 695 & 78.07 \\
        \midrule
        \multicolumn{2}{l|}{\textbf{S: PSPNet-R18}} & \centering \multirow{6}{*}{12.61} & \centering \multirow{6}{*}{109} & 72.55 \\
        SKD~\cite{liu2019structured} & \textcolor{gray}{CVPR'19} & & & 73.29 \\
        IFVD~\cite{wang2020intra} & \textcolor{gray}{ECCV'20} & & & 73.71 \\
        CWD~\cite{shu2021channel} & \textcolor{gray}{ICCV'21} & & & 74.36 \\
        CIRKD~\cite{yang2022cross} & \textcolor{gray}{CVPR'22} & & & 74.73 \\
        MasKD~\cite{huang2023masked} & \textcolor{gray}{ICLR'23} & & & 75.34 \\
        \rowcolor{gray!20} \textbf{Ours} & & & & \textbf{75.99} \\
        \bottomrule
    \end{tabular}
    }
    \centering
    \caption{Comparison of knowledge distillation methods for semantic segmentation on Cityscapes with the PSPNet-R18 student model. T: teacher model, S: student baseline model.}
    \label{tab:pspnet_r18}
\end{table}

\subsection{Runtime Efficiency Analysis}
\begin{table}[ht]
    \centering
    \begin{tabular}{p{1.25cm}| p{1.2cm} p{1.2cm} p{1.2cm} p{1.2cm}}
        \toprule
        \textbf{Model} & \textbf{Params (M)} & \textbf{FLOPs (G)} & \textbf{Mem (MB)} & \textbf{FPS} \\
        \midrule
        \multicolumn{5}{c}{\textit{Single-Stage Detectors (RetinaNet)}} \\
        \midrule
         T: X101  & 95.86  & 424  & 367  & 29.4   \\
         T: R101  & 56.96  & 283  & 220  & 30.7   \\
         S: R50   & 37.97  & 215  & 148  & 41.9   \\
        \midrule
        \multicolumn{5}{c}{\textit{Two-Stage Detectors (Faster R-CNN)}} \\
        \midrule
        T: X101  & 135.0  & 2014  & 528  & 20.6   \\
        T: R101  & 60.75  & 255  & 244  & 31.1   \\
        S: R50    & 41.75  & 187  & 171  & 42.1   \\
        \midrule
        \multicolumn{5}{c}{\textit{Anchor-Free Detectors (RepPoints)}} \\
        \midrule
        T: X101  & 94.74  & 380  & 230  & 16.6   \\
        T: R101  & 55.84  & 239  & 224  & 24.5   \\
        S: R50   & 36.85  & 171  & 151  & 31.4   \\
        \bottomrule
    \end{tabular}
    \caption{Efficiency comparison of different object detectors. In addition to the number of parameters (Params) and FLOPs, we report CUDA memory usage (Mem) and inference speed (FPS), measured on an NVIDIA A100 GPU (80GB). For two-stage detectors using X101 as the teacher backbone, Cascade Mask R-CNN is used. All models are evaluated with an input resolution of 1088×800. T: teacher model, S: student model.}
    \label{tab:efficiency}
\end{table}

In addition to the number of parameters (Params) and FLOPs, \cref{tab:efficiency} compares the runtime efficiency of different object detectors in terms of memory footprint and inference speed (frames per second, FPS). The large two-stage teacher model with X101 as its backbone incurs the highest computational cost, requiring more than 2000G FLOPs and consuming significant CUDA memory, making it less suitable for resource-constrained applications. In contrast, the student models with R50 backbones have much lower complexity while maintaining competitive performance. Inference speed is generally correlated with model size, with lighter models (e.g., RetinaNet-R50) achieving over 40 FPS, while heavier models with X101 backbones (e.g., RepPoints-X101) operate at only around 20 FPS.


Additionally, \cref{tab:cuda_fps_a100} presents hardware-aware efficiency metrics for segmentation models, including CUDA memory usage and FPS, complementing the earlier tables that include comparisons of parameter counts and FLOPs. The teacher model DeepLabV3-R101 has the highest memory consumption (1828 MB) while achieving only 11.9 FPS. In contrast, the DeepLabV3-R18 and PSPNet-R18 student models achieve over 59 FPS (nearly 6 times faster than the teacher) while consuming significantly less memory (568MB and 621MB, respectively), making them better suited for real-time applications. Notably, DeepLabV3-MBV2, despite having fewer FLOPs and parameters, runs at a lower FPS (52.9) than DeepLabV3-R18 (59.2). This is likely due to the inefficiency of depthwise separable convolutions on the specific CUDA-based GPU used for evaluation. While they reduce FLOPs, their lower computational density results in suboptimal parallelization, making them less efficient than standard dense convolutions, as used in ResNet-18.
\begin{table}[h!]
    \centering
    \begin{tabular}{lcc}
        \toprule
        \textbf{Model} & \textbf{Mem (MB)} & \textbf{FPS} \\
        \midrule
        T:DeepLabV3-R101  & 1828  & 11.9  \\
        S:DeepLabV3-R18   & 568   & 59.2  \\
        S:DeepLabV3-MBV2 & 470   & 52.9  \\
        S:PsPNet-R18      & 621   & 59.4  \\
        \bottomrule
    \end{tabular}
    \caption{Efficiency comparison of different segmentation models. CUDA memory usage and FPS for different segmentation models tested on an NVIDIA A100 GPU (80GB). Parameter counts and FLOPs are reported in earlier tables. All models are evaluated with an input resolution of 1088×800. T: teacher model, S: student model.}
    \label{tab:cuda_fps_a100}
\end{table}

\section{Ablation Studies} \label{sec:analysis}


\subsection{Spatial and Channel Masking}
\label{subsec:spatial_vs_channel}
\cref{tab:spatial_channel_study} demonstrates the impact of spatial and channel masking on knowledge distillation performance. With our student-teacher feature fusion module, both spatial-only and channel-only masking improve mAP over the baseline student (R50). Spatial masking achieves a slightly higher mAP gain (40.9 vs. 40.4) due to its preservation of spatial dimensions that is beneficial for dense prediction tasks. Meanwhile, combining both spatial and channel masking further improves performance, achieving the highest mAP of 41.2.
AP$_s$ benefits the most from spatial masking, while AP$_m$ and AP$_l$ see the greatest gains when both masking strategies are used together. These results highlight the complementary nature of spatial and channel masking, validating the design of our approach.
\begin{table}[h!]
\centering
\resizebox{\columnwidth}{!}{%
\begin{tabular}{lcccccc}
\toprule
\textbf{Method} & \textbf{mAP} & \textbf{AP$_{50}$} & \textbf{AP$_{75}$} & \textbf{AP$_s$} & \textbf{AP$_m$} & \textbf{AP$_l$} \\
\midrule
T: R101 & 38.9 & 58.0 & 41.5 & 21.0 & 42.8 & 52.4 \\
S: R50 & 37.4 & 56.7 & 39.6 & 20.0 & 40.7 & 49.7 \\
\midrule
Spatial (Sp)  & 40.9 & 60.4 & 43.4 & 25.4 & 44.3 & 52.3 \\
Channel (Ch) & 40.4 & 60.1 & 43.2 & 24.5 & 44.1 & 52.3 \\
\rowcolor{gray!20} \textbf{Ours (Sp+Ch)} & \textbf{41.2} & \textbf{60.6} & \textbf{44.1} & \textbf{24.6} & \textbf{45.5} & \textbf{54.1} \\
\bottomrule
\end{tabular}
}
\caption{Ablation study of spatial and channel masking strategies. Model: RetinaNet, Dataset: COCO. T: teacher model, S: student model.}
\label{tab:spatial_channel_study}
\end{table}

\subsection{Query Selection in Cross-Attention}
\label{subsec:query_selection}
\begin{table}[h!]
\centering
\resizebox{0.9\columnwidth}{!}{%
\begin{tabular}{lcccccc}
\toprule
\textbf{Method} & \textbf{mAP}  & \textbf{AP$_s$} & \textbf{AP$_m$} & \textbf{AP$_l$} \\
\midrule
Query from S & 41.0  & 24.4 & 45.1 & 54.0 \\
\rowcolor{gray!20} \textbf{Query from T (Ours)} & \textbf{41.2}  & \textbf{24.6} & \textbf{45.5} & \textbf{54.1} \\
\bottomrule
\end{tabular}
}
\caption{Comparison of query selection strategies in our student-teacher cross-attention feature fusion (STCA-FF) module. Model: RetinaNet, Dataset: COCO. T: teacher model, S: student model.}
\label{tab:query_selection}
\end{table}

\begin{table*}[h!]
\centering
\resizebox{1.35\columnwidth}{!}{%
\begin{tabular}{lcccccc}
\toprule
\textbf{Method} & \textbf{mAP} & \textbf{AP$_{50}$} & \textbf{AP$_{75}$} & \textbf{AP$_s$} & \textbf{AP$_m$} & \textbf{AP$_l$} \\
\midrule
T: R101 & 38.9 & 58.0 & 41.5 & 21.0 & 42.8 & 52.4 \\
S: R50 & 37.4 & 56.7 & 39.6 & 20.0 & 40.7 & 49.7 \\
No Masking & 37.4 & 57.1 & 40.0 & 20.8 & 40.8 & 50.9 \\
Fixed Masking from Teacher & 39.8 & 59.0 & 42.5 & 21.5 & 43.9 & 54.0 \\
Adaptive Masking from Teacher & 39.9 & 59.1 & 42.9 & 21.7 & 43.7 & 53.9 \\
\rowcolor{gray!20} \textbf{Our ACAM-KD} & \textbf{41.2} & \textbf{60.6} & \textbf{44.1} & \textbf{24.6} & \textbf{45.5} & \textbf{54.1} \\
\bottomrule
\end{tabular}
}
\caption{Ablation study of fixed and adaptive masking strategies for knowledge distillation. Model: RetinaNet, Dataset: COCO. T: teacher model, S: student model.}
\label{tab:ablation_study}
\end{table*}

\cref{tab:query_selection} compares different query selection strategies in our student-teacher cross-attention feature fusion (STCA-FF) module. Our default approach, in which the teacher generates the query while the student provides the key and value, achieves 41.2 mAP, outperforming the alternative configuration where the student defines the query (41.0 mAP). While both configurations yield similar results, the slight performance difference suggests that having the teacher generate the query enables better-guided knowledge transfer, as the teacher’s features provide a strong reference for the student to align with.

\subsection{Fixed and Adaptive Masking}
\label{subsec:fixed_vs_adaptive}

\cref{tab:ablation_study} compares fixed and adaptive masking strategies in knowledge distillation. The no-masking baseline achieves 37.4 mAP, indicating that the student model benefits only marginally from standard distillation without strategic feature selection. Fixed distillation masking learned by the teacher offline (as in \cite{huang2023masked}) improves mAP to 39.8. Allowing the teacher-driven masking to adapt during distillation (i.e., adaptive masking from the teacher) refines the selection process but results in only a slight improvement to 39.9 mAP. This suggests that making the mask learnable alone has limited impact when the knowledge source remains strictly from the teacher.
Our ACAM-KD, which combines student-teacher cross-attention feature fusion with adaptive masking, achieves the highest performance of 41.2 mAP. Notably, our method is particularly effective in handling small objects, as AP$_s$ shows the largest improvement, reaching 24.6. The superiority of ACAM-KD stems from the cooperative interaction between the teacher and student in identifying the most suitable knowledge for the student's evolving needs at different distillation stages.

\subsection{Effect of Diversity Loss} 
\begin{figure}[h]
    \centering
    \includegraphics[width=0.45\textwidth, trim={0in 0in 0in 0.75in},clip]{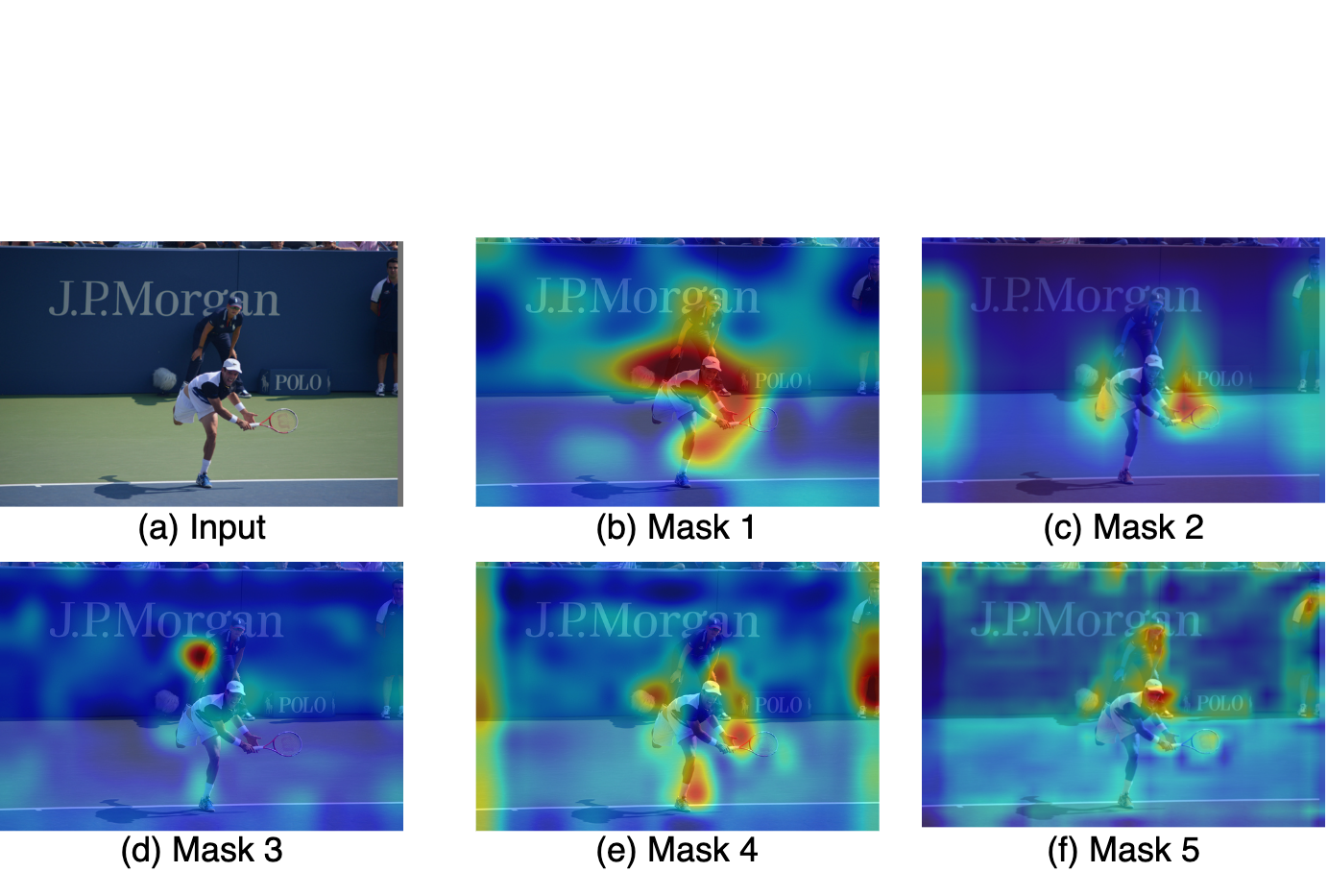}
\caption{
    Visualization of spatial attention masks associated with different learnable selection units. Variations in highlighted regions, encouraged by our diversity loss, ensure complementary feature learning for effective knowledge distillation. Warmer colors (red/yellow) indicate higher attention, while cooler colors (blue) denote lower attention.
}\label{fig:mask_analysis_spatial}
\end{figure}

\cref{fig:mask_analysis_spatial} shows that different spatial masks focus on distinct regions, emphasizing their complementary roles in knowledge transfer. This variation is encouraged by the diversity loss, which prevents multiple masks from collapsing into similar patterns. By promoting diversity, the learned masks ensure broader feature coverage, enabling the student model to acquire more comprehensive and informative knowledge.

\section{Conclusion}
\label{sec:conclusion}

Existing feature-based knowledge distillation methods rely on static teacher-driven supervision, failing to adapt to the student’s evolving learning state. These approaches often neglect the importance of dynamic student-teacher interactions and joint spatial-channel feature selection. In this paper, we propose Adaptive and Cooperative Attention Masking for Knowledge Distillation (ACAM-KD), which integrates student-teacher cross-attention feature fusion and adaptive spatial-channel masking. Our method
grants the student greater autonomy and facilitates a more interactive and adaptive distillation process,
resulting in consistent performance gains in dense prediction tasks (e.g., object detection and semantic segmentation). 
The results highlight the effectiveness of dynamically adjusting knowledge transfer to the student's needs, providing a promising approach for efficient model compression.
{
    \small
    \bibliographystyle{ieeenat_fullname}
    \bibliography{main}

\begin{thebibliography}{28}
\providecommand{\natexlab}[1]{#1}
\providecommand{\url}[1]{\texttt{#1}}
\expandafter\ifx\csname urlstyle\endcsname\relax
  \providecommand{\doi}[1]{doi: #1}\else
  \providecommand{\doi}{doi: \begingroup \urlstyle{rm}\Url}\fi

\bibitem[Chen et~al.(2017)Chen, Choi, Yu, Han, and Chandraker]{chen2017learning}
Guobin Chen, Wongun Choi, Xiang Yu, Tony Han, and Manmohan Chandraker.
\newblock Learning efficient object detection models with knowledge distillation.
\newblock \emph{Advances in Neural Information Processing Systems}, 30, 2017.

\bibitem[Chen et~al.(2019)Chen, Wang, Pang, Cao, Xiong, Li, Sun, Feng, Liu, Xu, Zhang, Cheng, Zhu, Cheng, Zhao, Li, Lu, Zhu, Wu, Dai, Wang, Shi, Ouyang, Loy, and Lin]{mmdetection}
Kai Chen, Jiaqi Wang, Jiangmiao Pang, Yuhang Cao, Yu Xiong, Xiaoxiao Li, Shuyang Sun, Wansen Feng, Ziwei Liu, Jiarui Xu, Zheng Zhang, Dazhi Cheng, Chenchen Zhu, Tianheng Cheng, Qijie Zhao, Buyu Li, Xin Lu, Rui Zhu, Yue Wu, Jifeng Dai, Jingdong Wang, Jianping Shi, Wanli Ouyang, Chen~Change Loy, and Dahua Lin.
\newblock {MMDetection}: Open mmlab detection toolbox and benchmark.
\newblock \emph{arXiv preprint arXiv:1906.07155}, 2019.

\bibitem[Chen et~al.(2018)Chen, Zhu, Papandreou, Schroff, and Adam]{chen2018encoder}
Liang-Chieh Chen, Yukun Zhu, George Papandreou, Florian Schroff, and Hartwig Adam.
\newblock Encoder-decoder with atrous separable convolution for semantic image segmentation.
\newblock In \emph{Proceedings of the European conference on computer vision (ECCV)}, pages 801--818, 2018.

\bibitem[Contributors(2020)]{mmseg2020}
MMSegmentation Contributors.
\newblock {MMSegmentation}: Openmmlab semantic segmentation toolbox and benchmark.
\newblock \url{https://github.com/open-mmlab/mmsegmentation}, 2020.

\bibitem[Cordts et~al.(2016)Cordts, Omran, Ramos, Rehfeld, Enzweiler, Benenson, Franke, Roth, and Schiele]{cordts2016cityscapes}
Marius Cordts, Mohamed Omran, Sebastian Ramos, Timo Rehfeld, Markus Enzweiler, Rodrigo Benenson, Uwe Franke, Stefan Roth, and Bernt Schiele.
\newblock The cityscapes dataset for semantic urban scene understanding.
\newblock In \emph{Proceedings of the IEEE conference on computer vision and pattern recognition}, pages 3213--3223, 2016.

\bibitem[Dai et~al.(2021)Dai, Jiang, Wu, Bao, Wang, Liu, and Zhou]{dai2021general}
Xing Dai, Zeren Jiang, Zhao Wu, Yiping Bao, Zhicheng Wang, Si Liu, and Erjin Zhou.
\newblock General instance distillation for object detection.
\newblock In \emph{Proceedings of the IEEE/CVF Conference on Computer Vision and Pattern Recognition}, pages 7842--7851, 2021.

\bibitem[Du et~al.(2021)Du, Zhang, Chang, Liu, Chen, Chen, et~al.]{zhixing2021distilling}
Zhixing Du, Rui Zhang, Ming Chang, Shaoli Liu, Tianshi Chen, Yunji Chen, et~al.
\newblock Distilling object detectors with feature richness.
\newblock \emph{Advances in Neural Information Processing Systems}, 34:\penalty0 5213--5224, 2021.

\bibitem[Heo et~al.(2019)Heo, Kim, Yun, Park, Kwak, and Choi]{heo2019comprehensive}
Byeongho Heo, Jeesoo Kim, Sangdoo Yun, Hyojin Park, Nojun Kwak, and Jin~Young Choi.
\newblock A comprehensive overhaul of feature distillation.
\newblock In \emph{Proceedings of the IEEE/CVF international conference on computer vision}, pages 1921--1930, 2019.

\bibitem[Huang et~al.(2023{\natexlab{a}})Huang, Zhang, You, Wang, Qian, Cao, and Xu]{huang2023masked}
Tao Huang, Yuan Zhang, Shan You, Fei Wang, Chen Qian, Jian Cao, and Chang Xu.
\newblock Masked distillation with receptive tokens.
\newblock In \emph{Proceedings of the International Conference on Learning Representations (ICLR)}, 2023{\natexlab{a}}.

\bibitem[Huang et~al.(2023{\natexlab{b}})Huang, Zhang, Zheng, You, Wang, Qian, and Xu]{huang2023knowledge}
Tao Huang, Yuan Zhang, Mingkai Zheng, Shan You, Fei Wang, Chen Qian, and Chang Xu.
\newblock Knowledge diffusion for distillation.
\newblock \emph{Advances in Neural Information Processing Systems}, 36:\penalty0 65299--65316, 2023{\natexlab{b}}.

\bibitem[Li et~al.(2017)Li, Jin, and Yan]{li2017mimicking}
Quanquan Li, Shengying Jin, and Junjie Yan.
\newblock Mimicking very efficient network for object detection.
\newblock In \emph{Proceedings of the IEEE/CVF Conference on Computer Vision and Pattern Recognition}, pages 6356--6364, 2017.

\bibitem[Lin et~al.(2014)Lin, Maire, Belongie, Hays, Perona, Ramanan, Doll{\'a}r, and Zitnick]{lin2014microsoft}
Tsung-Yi Lin, Michael Maire, Serge Belongie, James Hays, Pietro Perona, Deva Ramanan, Piotr Doll{\'a}r, and C~Lawrence Zitnick.
\newblock Microsoft coco: Common objects in context.
\newblock In \emph{Proceedings of the European Conference on Computer Vision (ECCV)}, pages 740--755. Springer, 2014.

\bibitem[Lin et~al.(2017)Lin, Goyal, Girshick, He, and Doll{\'a}r]{lin2017focal}
Tsung-Yi Lin, Priya Goyal, Ross Girshick, Kaiming He, and Piotr Doll{\'a}r.
\newblock Focal loss for dense object detection.
\newblock In \emph{Proceedings of the IEEE/CVF International Conference on Computer Vision}, pages 2980--2988, 2017.

\bibitem[Liu et~al.(2019)Liu, Chen, Liu, Qin, Luo, and Wang]{liu2019structured}
Yifan Liu, Ke Chen, Chris Liu, Zengchang Qin, Zhenbo Luo, and Jingdong Wang.
\newblock Structured knowledge distillation for semantic segmentation.
\newblock In \emph{Proceedings of the IEEE/CVF conference on computer vision and pattern recognition}, pages 2604--2613, 2019.

\bibitem[Ren et~al.(2017)Ren, He, Girshick, and Sun]{Ren_2017}
Shaoqing Ren, Kaiming He, Ross Girshick, and Jian Sun.
\newblock Faster r-cnn: Towards real-time object detection with region proposal networks.
\newblock \emph{IEEE Transactions on Pattern Analysis and Machine Intelligence}, 2017.

\bibitem[Romero et~al.(2014)Romero, Ballas, Kahou, Chassang, Gatta, and Bengio]{romero2014fitnets}
Adriana Romero, Nicolas Ballas, Samira~Ebrahimi Kahou, Antoine Chassang, Carlo Gatta, and Yoshua Bengio.
\newblock Fitnets: Hints for thin deep nets.
\newblock \emph{arXiv preprint arXiv:1412.6550}, 2014.

\bibitem[Sandler et~al.(2018)Sandler, Howard, Zhu, Zhmoginov, and Chen]{sandler2018mobilenetv2}
Mark Sandler, Andrew Howard, Menglong Zhu, Andrey Zhmoginov, and Liang-Chieh Chen.
\newblock Mobilenetv2: Inverted residuals and linear bottlenecks.
\newblock In \emph{Proceedings of the IEEE conference on computer vision and pattern recognition}, pages 4510--4520, 2018.

\bibitem[Shu et~al.(2021)Shu, Liu, Gao, Yan, and Shen]{shu2021channel}
Changyong Shu, Yifan Liu, Jianfei Gao, Zheng Yan, and Chunhua Shen.
\newblock Channel-wise knowledge distillation for dense prediction.
\newblock In \emph{Proceedings of the IEEE/CVF international conference on computer vision}, pages 5311--5320, 2021.

\bibitem[Wang et~al.(2019)Wang, Yuan, Zhang, and Feng]{wang2019distilling}
Tao Wang, Li Yuan, Xiaopeng Zhang, and Jiashi Feng.
\newblock Distilling object detectors with fine-grained feature imitation.
\newblock In \emph{Proceedings of the IEEE/CVF Conference on Computer Vision and Pattern Recognition}, pages 4933--4942, 2019.

\bibitem[Wang et~al.(2020)Wang, Zhou, Jiang, Bai, and Xu]{wang2020intra}
Yukang Wang, Wei Zhou, Tao Jiang, Xiang Bai, and Yongchao Xu.
\newblock Intra-class feature variation distillation for semantic segmentation.
\newblock In \emph{Computer Vision--ECCV 2020: 16th European Conference, Glasgow, UK, August 23--28, 2020, Proceedings, Part VII 16}, pages 346--362. Springer, 2020.

\bibitem[Yang et~al.(2022{\natexlab{a}})Yang, Ochal, Storkey, and Crowley]{yang2022prediction}
Chenhongyi Yang, Mateusz Ochal, Amos Storkey, and Elliot~J Crowley.
\newblock Prediction-guided distillation for dense object detection.
\newblock In \emph{European Conference on Computer Vision}, pages 123--138. Springer, 2022{\natexlab{a}}.

\bibitem[Yang et~al.(2022{\natexlab{b}})Yang, Zhou, An, Jiang, Xu, and Zhang]{yang2022cross}
Chuanguang Yang, Helong Zhou, Zhulin An, Xue Jiang, Yongjun Xu, and Qian Zhang.
\newblock Cross-image relational knowledge distillation for semantic segmentation.
\newblock In \emph{Proceedings of the IEEE/CVF conference on computer vision and pattern recognition}, pages 12319--12328, 2022{\natexlab{b}}.

\bibitem[Yang et~al.(2019)Yang, Liu, Hu, Wang, and Lin]{yang2019reppoints}
Ze Yang, Shaohui Liu, Han Hu, Liwei Wang, and Stephen Lin.
\newblock Reppoints: Point set representation for object detection.
\newblock In \emph{The IEEE International Conference on Computer Vision (ICCV)}, 2019.

\bibitem[Yang et~al.(2022{\natexlab{c}})Yang, Li, Jiang, Gong, Yuan, Zhao, and Yuan]{yang2022focal}
Zhendong Yang, Zhe Li, Xiaohu Jiang, Yuan Gong, Zehuan Yuan, Danpei Zhao, and Chun Yuan.
\newblock Focal and global knowledge distillation for detectors.
\newblock In \emph{Proceedings of the IEEE/CVF Conference on Computer Vision and Pattern Recognition}, pages 4643--4652, 2022{\natexlab{c}}.

\bibitem[Yang et~al.(2022{\natexlab{d}})Yang, Li, Shao, Shi, Yuan, and Yuan]{yang2022masked}
Zhendong Yang, Zhe Li, Mingqi Shao, Dachuan Shi, Zehuan Yuan, and Chun Yuan.
\newblock Masked generative distillation.
\newblock In \emph{European conference on computer vision}, pages 53--69. Springer, 2022{\natexlab{d}}.

\bibitem[Zhang and Ma(2020)]{zhang2020improve}
Linfeng Zhang and Kaisheng Ma.
\newblock Improve object detection with feature-based knowledge distillation: Towards accurate and efficient detectors.
\newblock In \emph{International conference on learning representations}, 2020.

\bibitem[Zhang et~al.(2024)Zhang, Huang, Liu, Jiang, Cheng, and Zhang]{zhang2024freekd}
Yuan Zhang, Tao Huang, Jiaming Liu, Tao Jiang, Kuan Cheng, and Shanghang Zhang.
\newblock Freekd: Knowledge distillation via semantic frequency prompt.
\newblock In \emph{Proceedings of the IEEE/CVF Conference on Computer Vision and Pattern Recognition}, pages 15931--15940, 2024.

\bibitem[Zhao et~al.(2017)Zhao, Shi, Qi, Wang, and Jia]{zhao2017pyramid}
Hengshuang Zhao, Jianping Shi, Xiaojuan Qi, Xiaogang Wang, and Jiaya Jia.
\newblock Pyramid scene parsing network.
\newblock In \emph{Proceedings of the IEEE conference on computer vision and pattern recognition}, pages 2881--2890, 2017.

\end{thebibliography}
}

\end{document}